%% file: main.tex
\documentclass[letterpaper, 10pt, conference]{IEEEtran}
\usepackage{times}
\usepackage{xr}
\usepackage{amsmath,amssymb,mathrsfs,bm,gensymb}
\usepackage[]{algorithm2e}
\usepackage{graphicx}
\usepackage{xcolor}
\usepackage{wrapfig}
\usepackage{textcomp}

\definecolor{control}{RGB}{203, 65, 107}
\definecolor{shape}{RGB}{61, 153, 115}

\usepackage[sort,numbers]{natbib}
\usepackage{multicol}
\usepackage[bookmarks=true, hidelinks]{hyperref}

\begin{document}

\title{Reinforcement learning for freeform robot design}

\author{Author Names Omitted for Anonymous Review. Paper-ID [add your ID here]}


\author{
\IEEEauthorblockN{%
Muhan Li,
David Matthews,
Sam Kriegman}
\IEEEauthorblockA{
Northwestern University
}
}


\maketitle

\input{00_abstract}

\IEEEpeerreviewmaketitle

\begin{figure*}
    \centering
    \includegraphics[width=0.975\linewidth]{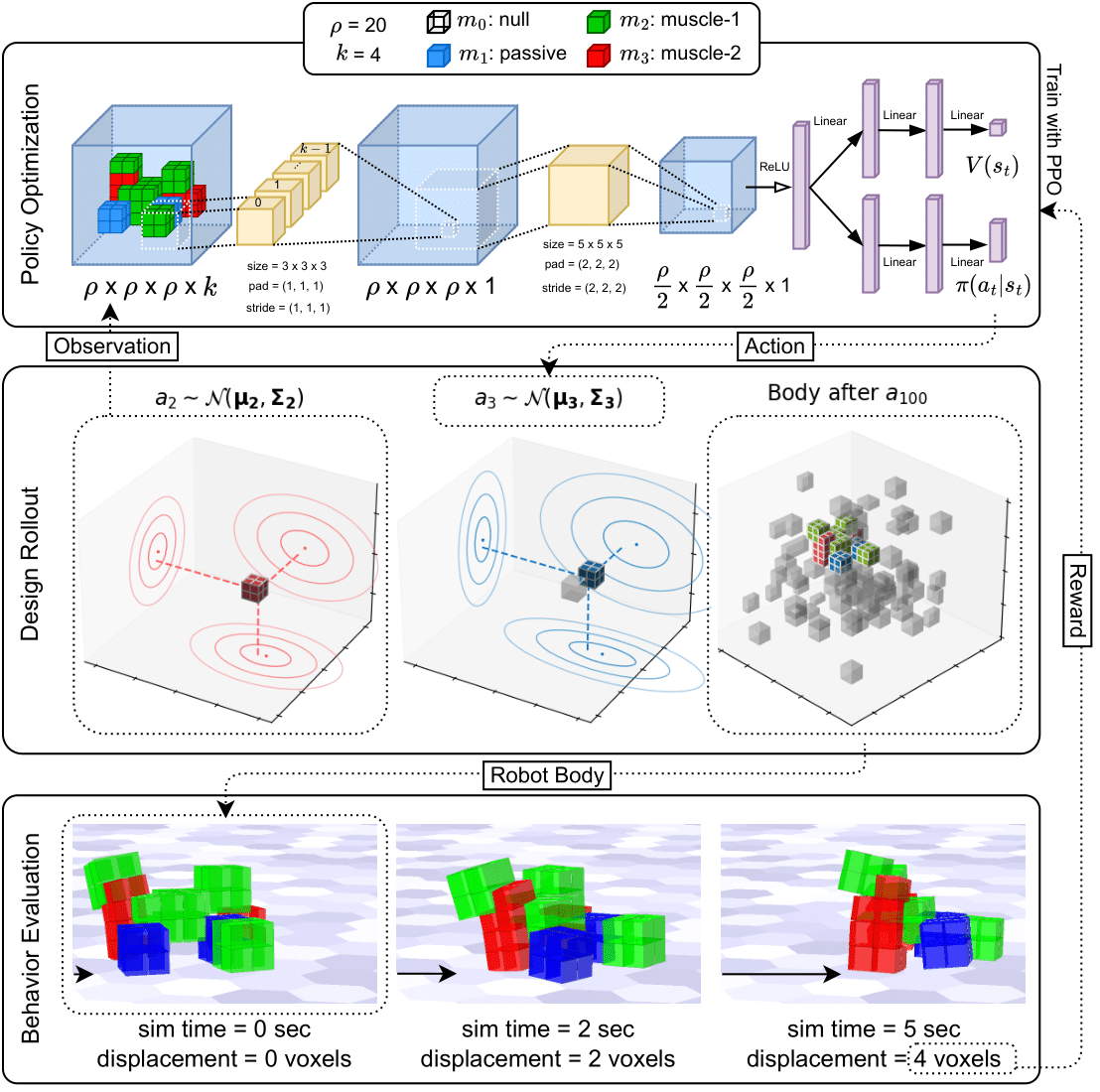}
    \caption{\textbf{Freeform robot design.}
    A policy (\textit{top row}) was trained to design increasingly motile robots,
    and predict their locomotive ability (critic; $V(s)$), 
    using a sequence of 100 design actions that freely position, overwrite or remove bundles of
    muscles (green and red) and
    passive tissue (dark blue) to form a robot (\textit{middle}) whose behavior in simulation (\textit{bottom}) determines the policy's reward 
    (\href{https://youtu.be/ybaEVDGvkTE}{\color{blue}\texttt{\textbf{youtu.be/ybaEVDGvkTE}}}).
    }
    \label{fig:summary}
\end{figure*}

\input{01_introduction}

\begin{figure*}
    \centering
    \includegraphics[width=\linewidth]{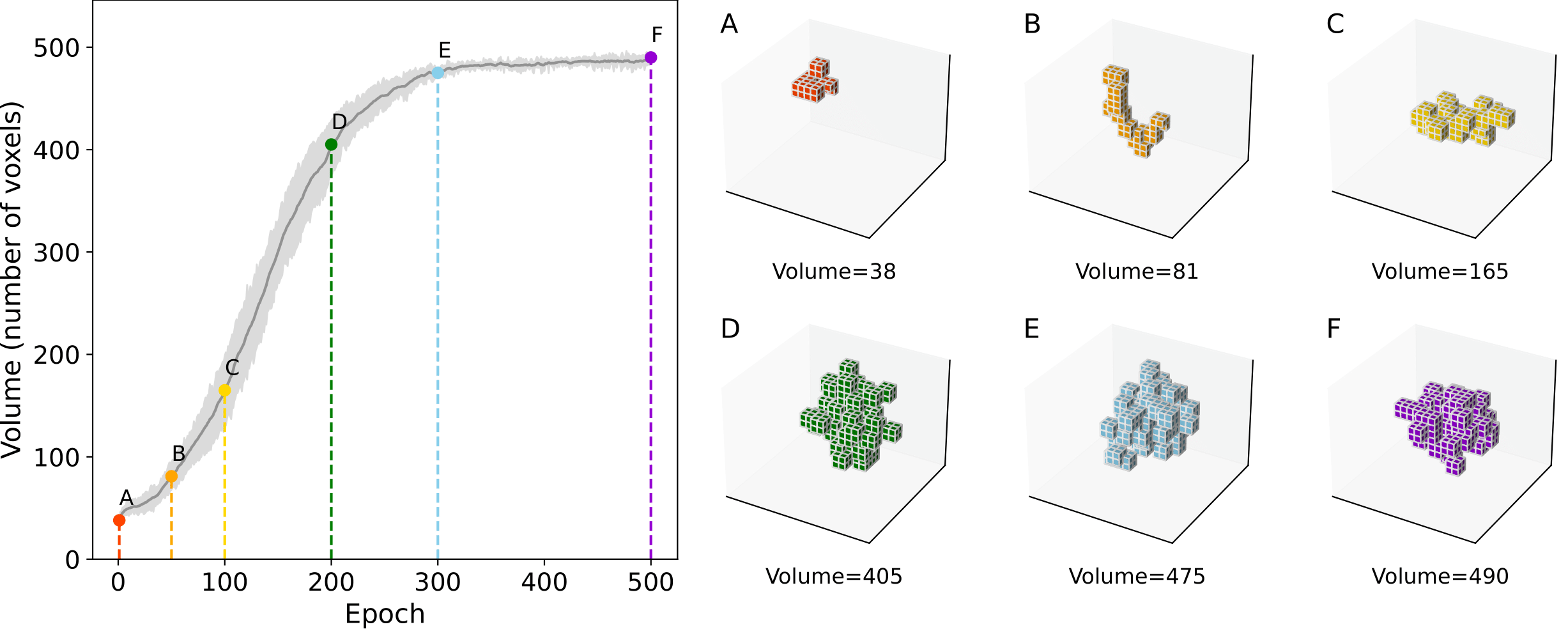}
    \caption{\textbf{Learning to design large nonparametric bodies.}
    \textit{Left:} Mean reward (volume; dark gray curve) and its $99\%$ Normal confidence interval (light gray bands) 
    across 5 independent learning trials.
    \textit{Right:} six bodies \mbox{(\textbf{A-F})} sampled along the reward curve from least to most voluminous.
    }
    \label{fig:volume}
\end{figure*}

\begin{figure*}
    \centering
    \includegraphics[width=\linewidth]{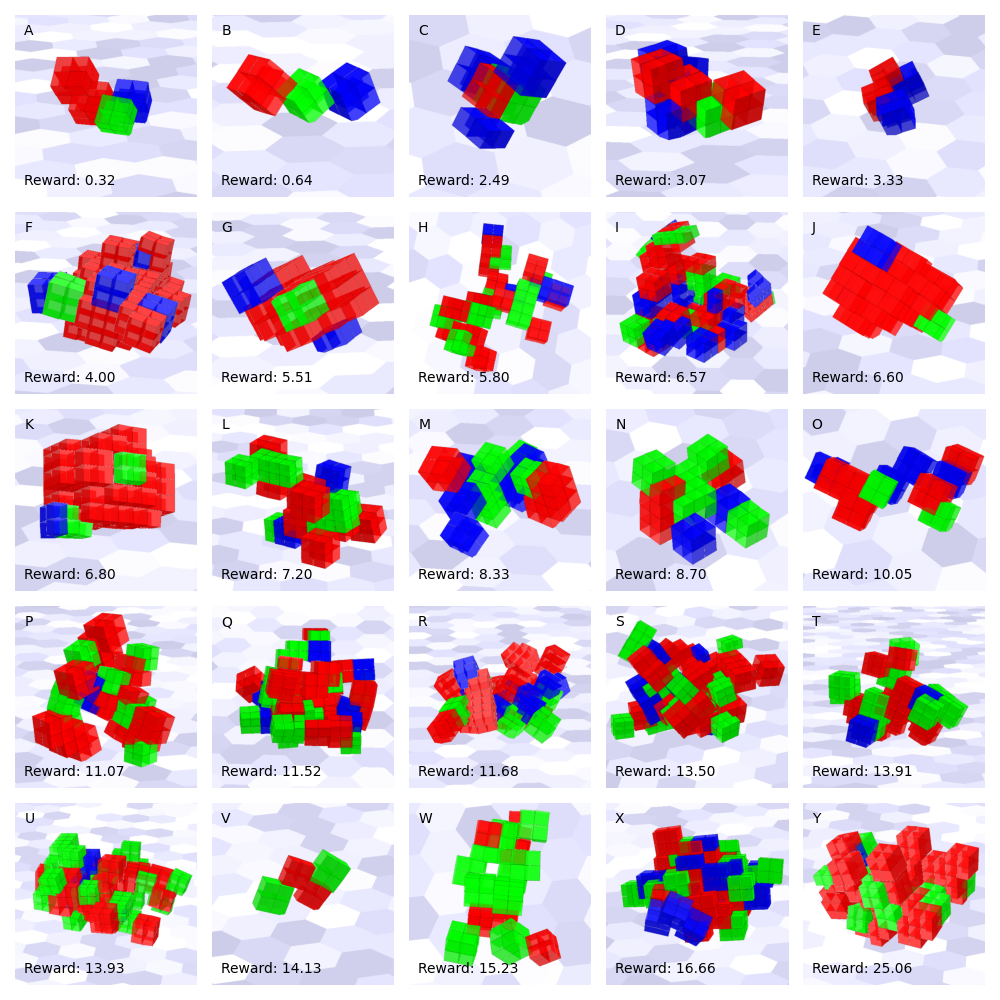}
    \vspace{-2em}
    \caption{\textbf{Designs for locomotion} 
    sampled across 5 independent trials at different epochs across training.
    Reward is net displacement (in voxel lengths) measured from evaluation start to end.
    }
    \label{fig:robot25}
\end{figure*}

\begin{figure*}
    \centering
    \includegraphics[width=\linewidth]{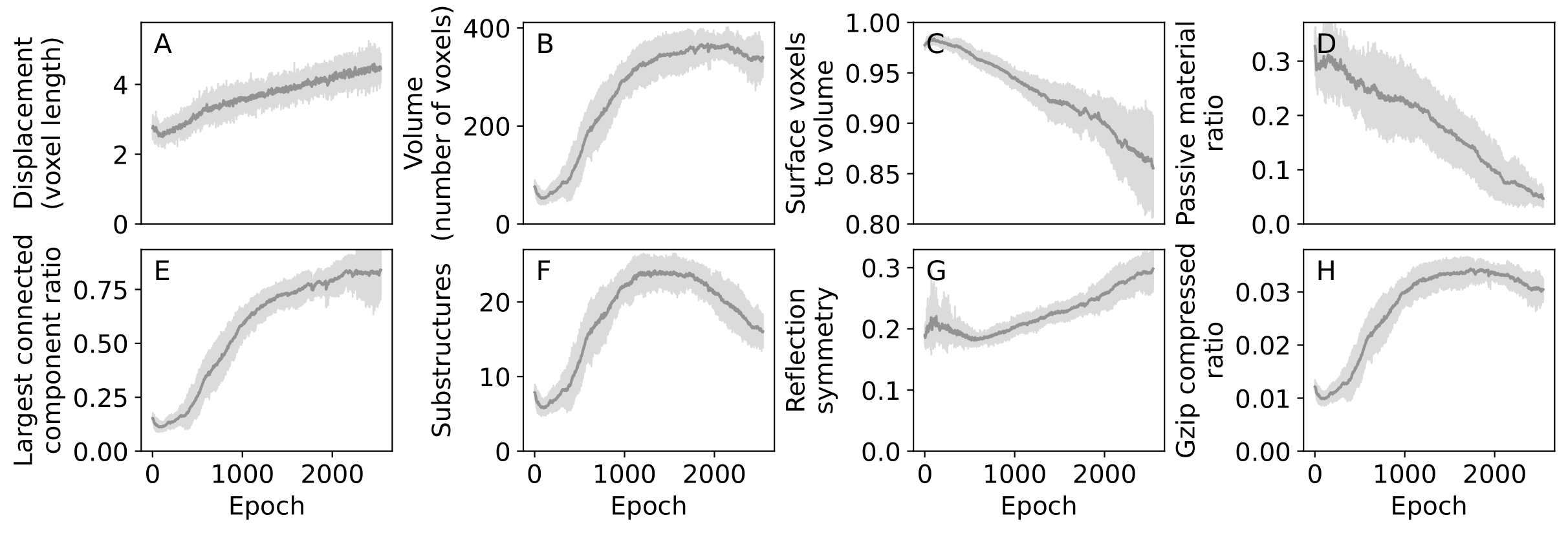}
    \caption{\textbf{Learning to design freeform robots.}
    Mean (dark gray curve) and $99\%$ Normal confidence intervals (light gray bands) 
    of reward (net displacement in voxel lengths; \textbf{A}),
    body volume (number of voxels; \textbf{B}),
    surface voxels to volume ratio (\textbf{C}),
    passive material ratio (\textbf{D}),
    largest connected component ratio (\textbf{E}),
    number of substructures (separate material regions; \textbf{F}),
    reflection symmetry (\textbf{G}),
    and compressability (using gzip; \textbf{H})
    during policy optimization across 5 independent learning trials. The policy learned to produce larger, more symmetrical bodies with less passive tissue with higher complexity as measured by the number of substructures and compression score.
    }
    \label{fig:locomotion}
\end{figure*}

\input{02_methods}


\input{03_results}

\begin{figure}[]
    \centering
    \includegraphics[width=0.6\linewidth]{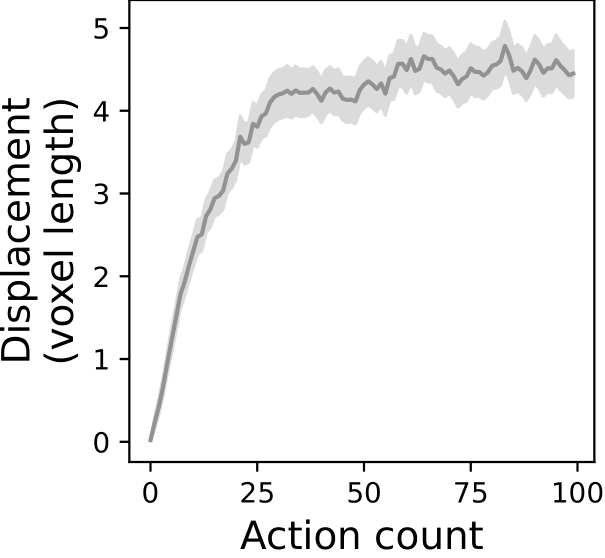}
    \caption{\textbf{Robustness.}
        Robots sampled from the final epoch of training were 
  simulated at each intermediate design step $t=1,2\ldots 99$.
  Behavioral reward of robots appears to converge within the first 30 actions.
  \label{fig:action_truncation}}
\end{figure}

\input{04_discussion}

\begin{table}[]
\centering
\caption{Hyperparameters used for \textbf{Vol}ume and \textbf{Loco}motion tasks.}
\begin{tabular}{l|l}
\hline
Hyperparameter                   & Value                                                                                  \\ \hline
CNN Size                         & (3x3x3, 4 kernels) (5x5x5, 1 kernel)                                                   \\
Actor MLP Size                   & Vol: (128, 128) / Loco: (256, 256)                                            \\
Critic MLP Size                  & Vol: (128, 128) / Loco: (256, 256)                                           \\
PPO Learning Rate                & 1e-4                                                                                   \\
PPO Batch Size                   & Vol: 25600 / Loco: 12800                                                      \\
PPO SGD Minibatch Size           & 128                                                                                    \\
PPO SGD iter per Minibatch      & Vol: 50 / Loco: 10                                                            \\
PPO Training Epochs              & Vol: 500 / Loco: 2500                                                         \\
PPO Discount Factor              & 0.99                                                                                   \\
PPO Critic Parameter Clipping    & Disabled (by setting to 1e5)                                                           \\
PPO GAE                          & Disabled                                                                               \\ \hline
\end{tabular}

\label{tab:hyperparam}
\end{table}


\input{05_ack}


\bibliographystyle{plainnat}
\bibliography{main}

\end{document}

%% file: 00_abstract.tex
\begin{abstract}
Inspired by the necessity of morphological adaptation in animals, a growing body of work has attempted to expand robot training to encompass physical aspects of a robot’s design. However, reinforcement learning methods capable of optimizing the 3D morphology of a robot have been restricted to reorienting or resizing the limbs of a predetermined and static topological genus. Here we show policy gradients for designing freeform robots with arbitrary external and internal structure. This is achieved through actions that deposit or remove bundles of atomic building blocks to form higher-level nonparametric macrostructures such as appendages, organs and cavities. Although results are provided for open loop control only, we discuss how this method could be adapted for closed loop control and sim2real transfer to physical machines in future.
\end{abstract}

%% file: 01_introduction.tex
\section{Introduction}
\label{sec:introducion}

Usually, control policies are optimized on a robot or embodied agent with a hand designed physical form. 
However, it has been shown that a robot's physical structure can facilitate or 
obstruct
policy optimization 
\cite{bongard2010utility,cheney2018scalable,iii2021taskagnostic,gupta2021embodied}.
This suggests that it may be useful to expand search to encompass the physical parameters of an agent as well, searching for structures that increasingly ease policy training.

Inspired by the evolution of animals, the automatic design of robots has been primarily achieved using evolutionary algorithms
\cite{sims1994competition,lipson2000automatic,hiller2012automatic,cheney2018scalable,kriegman2020xenobots,iii2021taskagnostic,gupta2021embodied}.
Although formal equivalences have been shown between learning and evolution \cite{watson2016can}, 
evolutionary robotics
relies entirely on random phylogenetic ``actions'' (mutations) to modify the robot's design, without any bias toward favorable outcomes (beneficial mutations).
That is, mutations at the beginning and end of evolution followed the same distribution and were not conditional on the state of the robot (its phenotype).
Evolutionary and learning algorithms differ also in the kinds of problems they have successfully solved; and,
until now, no 
reinforcement
learning algorithm has been shown to be capable of freeform robot design.

Reinforcement learning (RL) 
and RL-adjacent 
methods
have been used to 
lengthen or truncate
a segmented torso 
and its
pairs of jointed legs
\cite{zhao2020robogrammar};
resize a quadruped's four legs \cite{schaff2019jointly,ha2019reinforcement,ha2017joint} or a hexapod's six \cite{spielberg2017functional};
dynamically reorganize six modular limbs during behavior
\cite{pathak2019learning};
add or remove limbs at eight predefined locations \cite{Schaff-RSS-22};
and extrude four limbs from a torso that branch as unbalanced binary trees of depth three
\cite{yuan2022transformact}.
But, in every case,
the 
basic geometric shape
and 
internal structure
of the 
robot's 
torso and 
limbs
were 
predetermined and could not change during optimization.
Also, these algorithms could not alter the robot's topology (number of voids).

Voids and pores could be useful for robots as they
allow for 
internal carrying of objects, 
increase the robot's strength to weight ratio, 
and add surfaces for catalysis and heat exchange.
Our approach uses thousands of 
atomic elements---voxels---as building blocks of macrostructures,
which allows the number, placement, and 3D shape of limbs and voids to be optimized, simultaneously, with minimal assumptions.

Spherical particles \cite{bongard2003evolving} 
and cubic cells \cite{kriegman2021kinematic}
have been used in previous studies
to automatically optimize robot geometry and topology in a nonparametric manner, 
but they have yet to be leveraged 
for this purpose by RL methods.
This gap in the literature is due in part to the \textit{sui generis} nature of the design problem.
Freely adding, reshaping and removing body parts is fundamentally different than resizing limbs and re-weighting synapses.
The search landscape is much less forgiving of missteps:
Adding or removing even a single voxel along the underside of a bipedal walker's foot, for instance, could have catastrophic behavioral consequences.
Historically, this relegated the  problem 
to selection among random variations. 
But recent years have witnessed a sea change in robot design.

\citet{wang2023curriculumbased} used RL to design of 2D
agents
composed of 
49 
elastic
voxels.
However,
it was unclear if the agent's geometry varied during training, 
and if it did, if revisions were applied non-randomly.
When optimizing the agent for locomotion,
the agent's elasticity and motor layout 
were revised, 
but its square body shape remained unchanged.

\citet{ma2021diffaqua} also took a voxel-based approach.
Instead of RL,
gradient information from
differentiable simulation 
was used
to interpolate 
between 
human-selected basis shapes.
To ensure numerical stability,
these body shapes were required to share same topology.
The learner was thus trapped in the design space between hand-designed shapes and could not alter the robot's topology.
Also, the robots in \cite{ma2021diffaqua} were suspended in a liquid without contact modeling, which can be challenging to implement in a differentiable fashion, 
and without which land-based behaviors cannot be realized.
This is an important distinction for AI as land affords cognitive opportunities to agents that are absent in aquatic environments (e.g.~long-range vision and planning \cite{mugan2020spatial}).

\citet{matthews2023efficient} used differentiable simulation for freeform design and optimization of terrestrial yet 2D robots. 
The two dimensional design space was extruded into a third dimension yielding a physical 3D body plan.
More recently, 
others \cite{cochevelou2023differentiable,yuhn20234d}
have extended
differentiable design to
three dimensions.
While the robot from \cite{matthews2023efficient} was built and found to retain its behavior,
it is unclear how 
methods that rely on 
differentiable physics 
can proceed without 
a custom built
simulation
for every new task the robot faces.
RL, in contrast,
requires
only high-level reward definitions
and can thus design physical machines directly without recourse to simulations.

Thus, we here introduce a 
policy-gradient method 
for designing robots with freeform morphology.
The resulting robots are freeform in the sense that 
they can be generated 
with 
any 
3D shape
and
any internal 
3D organization of materials and voids, 
at any given cartesian resolution.

%% file: 02_methods.tex
\section{Methods}
\label{sec:methods}

\subsection{The environment.}
\label{sec:env}

Robots are here generated along a 
$\rho\times\rho\times\rho$
voxel grid $\mathbf{G}$,
yielding bodies with morphological resolution $\rho$.
We here set $\rho=20$.
Each voxel within a robot's body consists of a central point mass and up to six Euler-Bernoulli beams connecting to its neighbors (if any) on each face.
This allows for elastic twisting and stretching of one voxel relative to another.
Actuation is implemented by increasing and decreasing the rest length of a voxel's beams,
which produces volumetric expansion and contraction.
A Coulomb friction model is used for the surface plane.
For more details about the simulated physical environment, see \cite{hiller2014dynamic}.

\subsection{The state space.}
\label{sec:state}

The state space 
$s_t\in\{1,0\}^{\rho\times \rho\times \rho\times k}$
describes the robot's body plan: 
the 3D position $\mathbf{x}$ 
of each voxel in $\mathbf{G}$
and $k$ channels for 1-hot encoded material properties $\mathbf{m} = (m_0, m_1, \dots, m_{k-1})$, where $m_0$ is null material (empty space). 
If $\mathbf{m}$ is the zero vector then the null material is selected.
We here set $k=4$ 
with $m_1$ defined as passive tissue,
and $m_2$, $m_3$ defined as muscles that actuate in anti-phase at 4 Hz with amplitude $\pm10\%$.
Materials $m_1, m_2, m_3$ all have Young's modulus $10^5$ Pa (the stiffness of silicone rubber), density $1500\ kg/m^3$, Possion's ratio of 0.35, and coefficients of 1 and 0.5 for static and dynamic friction, respectively.

\subsection{The action space.}
\label{sec:act}

Each sampled action $\tilde{a}_t = (\mathbf{x}, \mathbf{m})$ 
deposits 
a bundle
of voxels $b_t$
centered at $\mathbf{x}$, 
with 
material properties $\mathbf{m}$.
We here set the bundle to be a $2\times2\times2$ cube of voxels.
Actions 
follow
a multivariate diagonal gaussian distribution 
$a_t \sim \mathcal{N}(\bm{\mu},\mathbf{\Sigma})$
with 
seven 
independent dimensions
corresponding to the 3D position and the 4D material properties of the bundle.
The support of the action distribution $a_t \in \mathbb{R}^7$ is mapped to that of $\mathbf{G}$ 
with clipping 
$f(a) = 1/2 + \text{clip}(a, -2, 2) / 4$,
which ensures that all bundles are centered inside of $\mathbf{G}$, but may ``hang over'' the edge.
If voxels of two or more bundles overlap, the material properties of the most recent action take priority.

\subsection{The reward.}
\label{sec:reward}

After a sequence of $T$ design actions, 
$\tilde{a}_1, \tilde{a}_2, \ldots, \tilde{a}_T$,
the largest contiguous collection of voxels is taken to be the robot's body, $\mathcal{B}$,
which is then evaluated
in a physics-based virtual environment \cite{hiller2014dynamic} 
and assigned a 
behavioral reward
equal to the net displacement of the robot.
The behavioral reward is only assigned to the reward of last step $r_T$ and zero is assigned to rewards of other steps $r_1, r_2, \ldots, r_{T-1}$.
Since rewarding for locomotion in terms of net displacement 
can create a local optimum of
ever taller morphologies that utilize falling as the main method of displacement to reach a higher reward,
we included a burn-in period in which 
the robot is given time to settle under gravity 
prior to recording the robot's initial location.

\subsection{The policy.}
\label{sec:policy}

The policy (Fig.~\ref{fig:summary}) was trained using PPO \cite{schulman2017proximal} and
comprises an actor $\pi(a_t|s_t)$, 
which takes design actions as described above in Sect.~\ref{sec:act},
and a design critic $V(s_t)$, which 
is trained to approximate the accumulated discounted reward at design step $t$:
\begin{equation}
    V(s_t) = \mathbb{E}_{\pi}\left[\,\sum_{t=0}^{T}\gamma^t r_t\right], \\[4pt]
\end{equation}
with discount factor $\gamma=0.99$.
The actor and critic share a base convolutional network but have independent fully connected layers,
since parameter sharing \cite{terry2020revisiting} has been shown to simplify training and speed convergence.
The base convolution network outputs the hidden embedding $h_t$ of state $s_t$. The policy distribution and the value function is thus defined as:
\begin{align}
\label{eq:policy}
    h_t &= \text{CNN}(s_t) \\[4pt]
    \bm{\mu} &= \text{MLP}_{\pi, \bm{\mu}}(h_t) \label{eq:policy_mu} \\[4pt]
    \mathbf{\Sigma} &= e^{\text{MLP}_{\pi, \mathbf{\Sigma}}(h_t)} \label{eq:policy_sigma} \\[4pt]
    \pi(a_t|s_t) &= \mathcal{N}(\bm{\mu},\mathbf{\Sigma}) \\[4pt]
    V(s_t) &= \text{MLP}_{V}(h_t)
\end{align}
The two MLPs: $\text{MLP}_{\pi, \bm{\mu}}$ and $\text{MLP}_{\pi, \mathbf{\Sigma}}(h_t)$ in Eq. \ref{eq:policy_mu} and \ref{eq:policy_sigma} were implemented as a single $\text{MLP}_{\pi}$.
The Xavier initialization method was used \cite{pmlr-v9-glorot10a} to ensure that the untrained 
policy
$\pi(a_t|s_t) \sim \mathcal{N}(\bm{\mu}=\mathbf{0},\mathbf{\Sigma}=\mathbf{1})$. 
When 
$a_t \sim \mathcal{N}(\bm{\mu}=\mathbf{0},\mathbf{\Sigma}=\mathbf{1})$ 
is mapped to voxel space 
through the clipping function $f(a)$, 
approximately $95\%$ of placed voxel bundle centers fall withing the grid $\textbf{G}$, 
while the rest fall onto the boundary. 
Hyperparameters are listed in Table \ref{tab:hyperparam}; unmentioned parameters used the default configuration provided in Version 2.0.1 of Ray \cite{liang2018rllib}.



%% file: 03_results.tex
\section{Results}
\label{sec:results}


We begin by quickly testing our design pipeline to ensure everything is working properly. 
To do so, we trained the policy for 500 epochs against a simple structural goal: 
maximize body volume using a single material (i.e.~$k=2$; Fig.~\ref{fig:volume}).
We repeated this experiment four times under different random seeds, yielding five independent trials (Fig.~\ref{fig:volume}).
The trained policy consistently learned to produce large bodies as evidenced by the significantly smaller bodies produced by the untrained policy 
($p < 0.01$).
The untrained policy tends to scatter voxel bundles throughout the workspace
with just 11.62\% 
of the voxels within the workspace, $\mathbf{G}$, contributing to the body $\mathcal{B}$, on average.
Over the course of training, the policy gradually learns to more efficiently utilize its actions such that, by the end of training, the vast majority ($82.62\%$) of voxels in $\mathbf{G}$ were also in $\mathcal{B}$.


Seeing that the policy was able to learn to design large coherent bodies we turned to the optimization of self-moving bodies: robots.
The reward was taken to 
be the maximum euclidean distance in the horizontal plane between the start and end position of each voxel in the robot, measured in voxel lengths,
across 
a 5 sec evaluation period (42K timesteps with stepsize 0.000118 sec),
which followed a 5 sec burn-in (42K timesteps with stepsize 0.000118 sec).
For this design task, we once again conducted five independent trials, each with their own unique random seed.
Each trial optimized a batch of 128 designs on a computer 
with either two A100 or three RTX A6000 GPUs 
for $\sim$28 hours. 

\begin{figure}[!b]
    \centering
    \includegraphics[width=\linewidth]{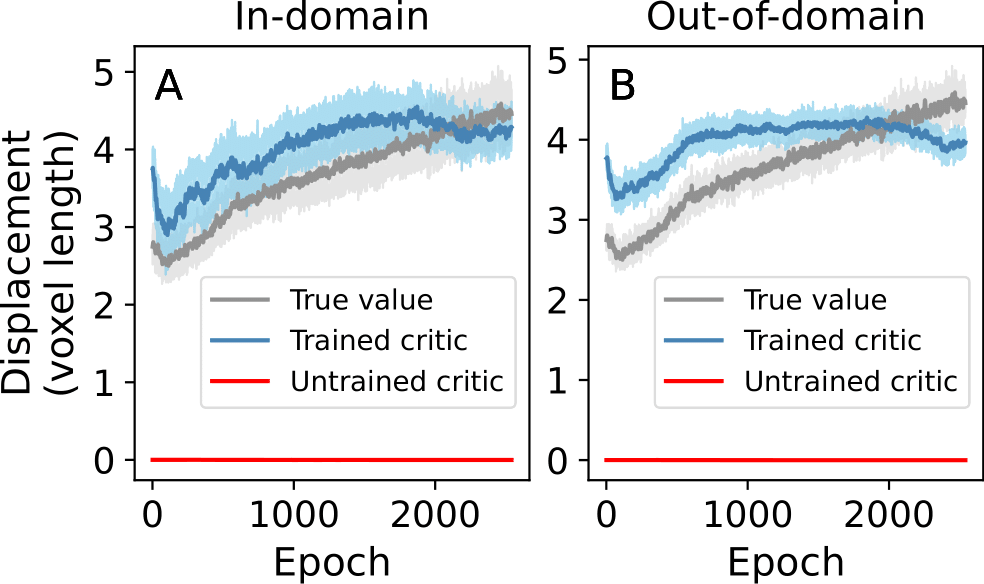}
    \caption{\textbf{Critic estimation of behavioral reward.}
    Predicted behavioral reward of the untrained (red) and trained (blue) critic against ground truth (gray) 
    for designs generated at each epoch during training. 
    Colored bands denote $99\%$ Normal confidence intervals across the 5 independent trials.
    The trained critic has learned the concept of locomotion as demonstrated by a significant improvement in prediction ability over the untrained critics.
    In-domain predictions are computed from bodies that each critic has seen during training (\textbf{A}), and out-of-domain predictions are computed from bodies taken from the sibling trials that each critic was not trained under (\textbf{B}). The trained critics generalize well to out-of-domain bodies, providing evidence that their understanding of the concept of locomotion extends beyond the specific bodies they saw during training.
    \label{fig:critic}}
\end{figure}

The policy, which was trained from scratch in each trial, 
learned to design better robots,
and the trained critic $V(s)$, which was also trained from scratch, 
learned the concept of locomotion,
as evidenced by 
the significantly less motile bodies produced by the untrained policy (Fig.~\ref{fig:locomotion}A; $p<0.01$)
and the significantly lower accuracy of the untrained critic (Fig.~\ref{fig:critic}; $p<0.01$), respectively.

A random sample of robots designed by the policy over the course of training can be seen in
Fig.~\ref{fig:robot25}.
The bodies with high rewards generally possess three design principles: 
(i) a relatively low passive material ratio (Fig.~\ref{fig:robot25}V,W,Y); 
(ii) multiple structures consisting of active materials in contact with the surface plane (Fig.~\ref{fig:robot25}U-Y); 
and 
(iii) a relatively high volume, which indirectly increases the number of active surface contacts (Fig.~\ref{fig:robot25}S,U,X,Y).
The model's discovery of these design principles is reflected in 
the body metrics shown in Fig.~\ref{fig:locomotion}.
The training generally 
increases the robot's body size (Fig.~\ref{fig:locomotion}B) while 
decreasing its surface area to volume ratio (Fig.~\ref{fig:locomotion}C),
and passive material ratio (Fig.~\ref{fig:locomotion}D).
This is accomplished by cohering the initially scattered design actions into a unified body (Fig.~\ref{fig:locomotion}E)
with more 
distinct substructures (Fig.~\ref{fig:locomotion}F), 
symmetry (Fig.~\ref{fig:locomotion}G),
and overall complexity as measured by the Gzip compression ratio (Fig.~\ref{fig:locomotion}H).

Reflection symmetry was calculated as the average ratio of voxels which remain unchanged when the body is mirrored across the three center planes of the robot's bounding box.

To determine the robustness of the optimized designs, 
we simulated the 
128 robots from the final batch 
in each independent trial,
this time simulating the robot after each design action instead of just after the last action $t=100$ (Fig. \ref{fig:action_truncation}).
The mean reward of these partially constructed robots quickly converges within the first 30 design actions, which
suggests that the additional 70 actions taken by the policy may be unnecessary.

%% file: 04_discussion.tex

\section{Discussion}
\label{sec:discussion}

In this paper, 
we showed that 
policy gradients
can, 
under certain conditions,
admit 
\textit{de novo} optimization
of 
nonparametric body plans.
However there were several limitations and assumptions that may be relaxed by future studies.
For example, 
the kind, size and shape of
building blocks made available to the policy,
as well as the coordinate system in which the policy acts,
were presupposed and held fixed throughout optimization.
Also,
reward was based solely on the behavior of the whole machine, 
without supplying reward signals \textit{during} body building.
Both of these limitations seem quite unnatural as
animal architectures 
are organized hierarchically
and take shape under 
rich and continual environmental feedback.
Indeed, while the employed design algorithm, reward function, and encoding (Fig.~\ref{fig:summary}) performed reasonably well for the structural goal we tested (volume; Fig.~\ref{fig:volume}), 
optimization was relatively slow for the tested behavioral goal (forward locomotion) as indicated by the average reward curve (Fig.~\ref{fig:locomotion}A) which had yet to converge after 28 hours.
But
the most important limitation of the
results is that they were provided for simulated robots only.

Future work will involve
transferring
simulated voxelized machines to physical ones 
\cite{hiller2012automatic,skylar2019voxelated,kriegman2020scalable}
using
sim2real methods 
that 
avoid 
difficult-to-simulate
structures \cite{koos2010crossing} 
and
dynamics \cite{jakobi1995noise}, 
or those that modify
the simulator itself
\cite{heiden2021neuralsim},
based on discrepancies between predicted (sim) and actual (real) behavior.
Simulator calibration, as from neural-augmentation of the underlying physics model \cite{heiden2021neuralsim}, could take place during morphological pretraining or it may be applied as a transferability filter just before finetuning.
The simulator would also be improved by
scaling the number of voxels \cite{aage2017giga}
and implementing
adaptive resolution grids to efficiently distribute these resources in space and time.
For example,
it may be 
beneficial to 
resolve
surface contact geometries
at a finer granularity
than internal materials, 
or to gradually increase global resolution over the course of training.
But this will ultimately
depend on the robot's intended niche.

Finally, it is important to note that while the process of generating a body was mediated by the environment, the behavior of the resulting robots was not.
Locomotion was instead coordinated by central pattern generators \cite{ijspeert2008central} and ``steered'' by the robot's overall shape and motor layout.
If sensor voxels 
are added to the design space,
the policy could
learn how to pattern receptors throughout the robot's body 
in order to capture relevant 
stimuli for transduction.
For instance,
macrostructures built out of
mechanoreceptive voxels \cite{truby2022fluidic}
could, 
depending on their arrangement, 
``listen to'' certain load signatures, and ignore others.
The emergence of
more
complex 
sensory organs 
and 
sense-guided behaviors 
could 
be 
hastened by
predator and prey scenarios~\cite{nolfi1998coevolving},
if they
trigger
a co-evolutionary arms race of morphological adaptations.

\section*{Code}
\label{sec:code}

The source code necessary to reproduce the results presented in this paper
can be found with MIT License on GitHub: 
\href{https://github.com/iffiX/RL4design}{\color{blue}\texttt{\textbf{github.com/iffiX/RL4design}}} 





%% file: 05_ack.tex
\section*{Acknowledgments}

We thank V.S.~Subrahmanian for lending computational resources that enabled this research, 
and Chris Fusting for helpful discussions.
This work was supported in part by 
the AI2050 program at Schmidt Sciences (Grant G-22-64506)
and
a seed grant from 
the Center for Engineering Sustainability and Resilience at Northwestern University.